\documentclass{article}

\usepackage[final,main]{Styles/neurips_2025}

\usepackage[T1]{fontenc}
\usepackage[utf8]{inputenc}
\usepackage{amsmath}
\usepackage{amssymb}
\usepackage{amsfonts}
\usepackage{url}
\usepackage{booktabs}
\usepackage{array}
\usepackage{graphicx}
\usepackage{float}
\usepackage{flafter}
\usepackage[section]{placeins}
\usepackage{hyperref}
\usepackage{microtype}

\hypersetup{
  colorlinks=true,
  linkcolor=blue,
  citecolor=blue,
  urlcolor=blue
}

\setcitestyle{numbers,square}

\begin{document}

\title{Adapting Multilingual Embedding Models to Turkish via\\Cross-Lingual Tokenizer Surgery and Offline Distillation}

\author{%
  M. Ali Bayram\thanks{Corresponding author.}\\
  Department of Computer Engineering\\
  Y{\i}ld{\i}z Technical University\\
  Istanbul, Turkey\\
  \texttt{malibayram20@gmail.com}
  \And
  Banu Diri\\
  Department of Computer Engineering\\
  Y{\i}ld{\i}z Technical University\\
  Istanbul, Turkey\\
  \texttt{diri@yildiz.edu.tr}
  \And
  Sava\c{s} Y{\i}ld{\i}r{\i}m\\
  Department of Computer Engineering\\
  Istanbul Bilgi University\\
  Istanbul, Turkey\\
  \texttt{savas.yildirim@bilgi.edu.tr}
}

\maketitle

\begin{abstract}
Sentence embeddings are a foundational component for semantic search, clustering, classification, and retrieval-augmented generation. This paper presents \emph{embeddingmagibu-200m}, a Turkish-focused sentence embedding model that produces 768-dimensional $\ell_2$-normalized vectors and supports an 8,192-token context window, far exceeding the 512-token limit of earlier BERT-based Turkish encoders. Instead of full pretraining, an efficient three-stage adaptation pipeline is introduced: (1)~construct a Turkish-optimized multilingual tokenizer with a $2^{17}=131{,}072$ vocabulary by pruning redundant tokens from the teacher's vocabulary and incorporating multilingual tokens via frequency analysis on a 40-language corpus, (2)~clone a teacher embedding model while preserving transformer backbone weights and initializing a compatible embedding table for the new vocabulary via mean-composition token mapping, and (3)~perform offline embedding distillation from precomputed teacher vectors using a cosine similarity objective over a balanced 40-language Wikipedia corpus. The resulting student model contains approximately 200M parameters and trains in roughly four hours on a single GPU by avoiding online teacher inference during training, at a total cost of \$5--\$20. Empirically, Pearson/Spearman correlations of 77.55\%/77.45\% are obtained on STSbTR, surpassing the 300M-parameter teacher model (73.84\%/72.92\%). On TR-MTEB (26 tasks), a mean score of 63.9\% is achieved (7th out of 26 models), providing a competitive cost--quality trade-off with 33\% fewer parameters than the teacher. To facilitate reproducibility and downstream use, all artifacts are released including model weights, tokenizer files, precomputed embedding datasets, and open-source cloning and distillation tooling.
\end{abstract}

\section{Introduction}
\label{sec:introduction}

Dense text embeddings have become foundational for modern NLP applications including semantic search, document clustering, duplicate detection, and retrieval-augmented generation~\cite{sbert_arxiv, mteb_arxiv}. Recent multilingual models such as Multilingual E5 and EmbeddingGemma~\cite{e5_arxiv, embeddinggemma_arxiv} deliver high general performance, but they must allocate capacity across many languages and carry large vocabularies, which can be suboptimal for a single language like Turkish. Recent analyses also show that tokenizer design materially affects downstream behaviour, motivating language-specific adaptation for morphologically rich languages~\cite{tokenizer_effects_arxiv}.

Turkish, as an agglutinative language with rich morphology, presents a particular challenge for standard subword tokenization algorithms. General multilingual tokenizers such as those of Gemma or LLaMA frequently fragment Turkish words into semantically meaningless subwords. For instance, a Turkish word with multiple suffixes such as \emph{evlerimizden} (``from our houses'') can be split into arbitrary, non-morphic subwords, degrading downstream semantic representation quality and expanding the token footprint. This footprint expansion directly reduces the effective context window and increases the $O(N^2)$ cost of self-attention. For Turkish, deployment also often requires both strong monolingual performance and an extended context window for document-level retrieval and indexing. Existing options are either large multilingual models or older monolingual BERT variants with limited 512-token context windows~\cite{sbert_arxiv}.

This paper introduces \emph{embeddingmagibu-200m}, a Turkish-focused sentence embedding model in the SentenceTransformers format\footnote{\url{https://www.sbert.net/}} with a maximum sequence length of 8,192 tokens and 768-dimensional normalized outputs. The model is designed to be efficient in parameter count (approximately 200M) while providing a long context window suitable for document-level retrieval. Rather than training from scratch, a high-capacity multilingual teacher is adapted to Turkish using an efficient three-stage pipeline:

\begin{enumerate}
    \item A Turkish-optimized multilingual tokenizer with a $2^{17}=131{,}072$ vocabulary is constructed. First, the $64\text{K}$ most frequent Turkish tokens are extracted from a tokenizer trained on the Cosmos Turkish Corpus\footnote{\url{https://huggingface.co/datasets/ytu-ce-cosmos/Cosmos-Turkish-Corpus-v1.0}}, and redundant/alternative tokens in the original teacher tokenizer are pruned. Then, a frequency analysis is performed on the Wikipedia-40-langs dataset\footnote{\url{https://huggingface.co/datasets/alibayram/wikipedia-40-langs}} to select multilingual tokens of varying lengths, yielding a 128K vocabulary that balances Turkish morphological alignment with robust multilingual capability.
    \item The teacher is cloned while preserving the transformer backbone weights and a compatible token embedding table is initialized for the new vocabulary via mean-composition token mapping. This procedure preserves the teacher's semantic space while adapting to the new vocabulary.
    \item Offline embedding distillation is performed by matching precomputed teacher vectors for approximately 580K examples drawn from a balanced 40-language Wikipedia corpus, using a cosine similarity objective.
\end{enumerate}

A key challenge addressed by this pipeline is that changing the tokenizer fundamentally alters the vocabulary size and token identities, making the original token embedding table incompatible. The cloning procedure addresses this by mapping each new token to one or more teacher tokens and composing their embeddings, preserving semantic information while adapting to the new vocabulary.

An end-to-end packaging and evaluation ecosystem is provided to encourage practical use and comparison of Turkish embedding models: (i)~model releases on Hugging Face and Ollama for local deployment, (ii)~a Hugging Face Space providing a TR-MTEB benchmark results explorer, (iii)~a PyPI package for cloning Transformers models to new tokenizers (\texttt{transformer-cloner}), (iv)~a distillation training package (\texttt{distil-trainer}), and (v)~released artifacts including precomputed teacher embeddings and exported benchmark outputs.\footnote{All resources are publicly available: model weights (\url{https://huggingface.co/magibu/embeddingmagibu-200m}), Ollama (\url{https://ollama.com/alibayram/embeddingmagibu-200m}), TR-MTEB explorer (\url{https://huggingface.co/spaces/magibu/mteb-turkish}), cloning tool (\url{https://pypi.org/project/transformer-cloner/}), and distillation package (\url{https://pypi.org/project/distil-trainer/}).}

The paper is organized as follows. Section~\ref{sec:related_work} reviews related work on Turkish embeddings, tokenizer adaptation, and distillation. Section~\ref{sec:method} describes the tokenizer training, vocabulary transfer via weight-preserving cloning, and offline distillation objective. Sections~\ref{sec:experiments}--\ref{sec:ablations} present the evaluation setup (STSbTR and TR-MTEB), main results, and ablation analyses. Sections~\ref{sec:limitations}--\ref{sec:reproducibility} discuss limitations and reproducibility details, followed by conclusions in Section~\ref{sec:conclusion}.

\section{Related Work}
\label{sec:related_work}

\paragraph{Turkish Sentence Embeddings and Benchmarks.}
Monolingual representation learning in Turkish has traditionally relied on encoder-only architectures, primarily BERT variants such as BERTurk~\cite{sbert_arxiv}. While these models are effective for sentence-level tasks, they are constrained by a 512-token context window, limiting their utility in document-level retrieval scenarios. The Turkish Massive Text Embedding Benchmark (TR-MTEB)~\cite{trmteb_paper}\footnote{\url{https://github.com/selmanbaysan/mteb_tr}} establishes a comprehensive multi-task suite covering classification, clustering, semantic textual similarity (STS), retrieval, natural language inference (NLI), and bitext mining tasks. TurkEmbed~\cite{turkembed_arxiv} demonstrates the benefits of training on native Turkish NLI and STS datasets. TabiBERT~\cite{tabibert_arxiv} introduces scale to Turkish representation learning by training ModernBERT-based encoders on massive corpora, confirming that monolingual foundation models remain competitive against larger multilingual architectures on local benchmarks.

\paragraph{Tokenizer Adaptation for Morphologically Rich Languages.}
Agglutinative languages like Turkish present significant challenges for subword tokenizers (e.g., Byte Pair Encoding or WordPiece) trained on predominantly English or multilingual corpora. Complex word forms constructed by appending suffixes to a root (e.g., \emph{yap-abili-yor-uz-dur}, ``we are probably able to do [it]'') are often fragmented into arbitrary, non-morphic subwords by general multilingual tokenizers. This fragmentation degrades representation quality by destroying morphological boundaries and expanding the per-sentence token count~\cite{morphscore_arxiv}. Recent studies show that tokenizer design materially affects downstream model performance and computational efficiency~\cite{tokenizer_effects_arxiv}. Standard techniques to address this include extending or replacing tokenizers during continual pretraining~\cite{tokenizer_adaptation_arxiv} or using hybrid tokenization strategies that blend statistical subwords with linguistically motivated units~\cite{hybrid_tokenization_arxiv}.

\paragraph{Vocabulary Transfer: WECHSEL vs.\ Mean-Composition Mapping.}
When a tokenizer is replaced, the input embedding table of the neural model becomes incompatible due to changes in token identities. Reinitializing the embedding layer randomly degrades performance and requires extensive pretraining to realign the embeddings with the transformer backbone. WECHSEL~\cite{wechsel_naacl2022} addresses this by aligning a new target-language tokenizer's embeddings with a source model's embeddings via bilingual static word vectors (e.g., fastText). In contrast, the mean-composition mapping used in this work is completely self-contained and deterministic: for each target subword, the string surface form is encoded using the teacher tokenizer, and the new embedding is initialized as the uniform average of the corresponding teacher embeddings. This eliminates external alignment noise and makes the process computationally trivial while preserving the teacher's semantic space.

\paragraph{Embedding Distillation.}
Knowledge distillation~\cite{reimers2020_multilingual} is a common technique to transfer semantic capabilities from a large teacher to a smaller student. Reimers and Gurevych~\cite{reimers2020_multilingual} pioneered multilingual sentence-level distillation by training a student model to match a monolingual teacher's representations using parallel corpora. To reduce the computational overhead of running the teacher model during training, offline distillation approaches precompute and store teacher embeddings, allowing the student to train independently. Recent multi-task models such as M3-Embedding~\cite{m3_embedding_arxiv} and mGTE~\cite{mgte_arxiv} emphasize self-distillation and multi-stage pretraining to achieve robust, long-context representations. By combining tokenizer surgery with offline distillation, large multilingual models can be adapted to target languages at a fraction of the traditional computational cost.

\section{Method}
\label{sec:method}

\begin{figure}[H]
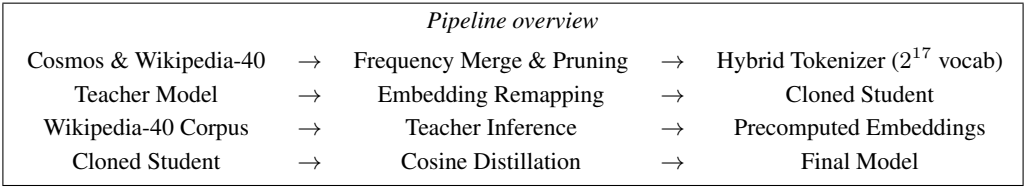

\centering
\fbox{\parbox{0.95\textwidth}{
\centering
\small
\emph{Pipeline overview}\\[0.5em]
\begin{tabular}{ccccc}
Cosmos \& Wikipedia-40 & $\rightarrow$ & Frequency Merge \& Pruning & $\rightarrow$ & Hybrid Tokenizer ($2^{17}$ vocab) \\[0.3em]
Teacher Model & $\rightarrow$ & Embedding Remapping & $\rightarrow$ & Cloned Student \\[0.3em]
Wikipedia-40 Corpus & $\rightarrow$ & Teacher Inference & $\rightarrow$ & Precomputed Embeddings \\[0.3em]
Cloned Student & $\rightarrow$ & Cosine Distillation & $\rightarrow$ & Final Model
\end{tabular}
}}
\caption{End-to-end pipeline for building \emph{embeddingmagibu-200m}. The teacher model is used only during embedding precomputation, enabling efficient offline distillation.}
\label{fig:pipeline}
\end{figure}

This section describes the end-to-end pipeline used to build \emph{embeddingmagibu-200m}: tokenizer training, model cloning with embedding remapping, teacher embedding precomputation, and embedding distillation. The pipeline is designed to retain the teacher's semantic space while reducing parameters via a Turkish-optimized vocabulary. Figure~\ref{fig:pipeline} illustrates the overall workflow.

\subsection{Tokenizer Construction}
\label{subsec:tokenizer_construction}

To optimize text representation for Turkish while preserving the multilingual capabilities of the teacher, we construct a custom hybrid tokenizer with a vocabulary size of $2^{17}=131{,}072$ tokens. This vocabulary size represents a balance between Turkish morphological coverage, representation capacity for other languages, and embedding table parameter footprint. The construction follows a structured multi-stage frequency analysis and pruning pipeline:

First, we perform frequency analysis on a tokenizer trained entirely on Turkish text using the Cosmos Turkish Corpus v1.0. From this tokenizer, we select the top $64\text{K}$ ($65{,}536$) most frequent Turkish tokens.
Using these selected Turkish tokens, we identify and prune alternative and redundant subwords or token sequences within the teacher model's original $256\text{K}$ tokenizer (from EmbeddingGemma-300M~\cite{embeddinggemma_arxiv}) that could be resolved or bypassed by these $64\text{K}$ Turkish tokens.
This step reduces lexical redundancy and ensures that Turkish text is encoded using the most morphologically natural and frequent Turkish subwords.

Second, to maintain the model's performance on other languages, we perform a frequency analysis on the Wikipedia-40-langs dataset. From this multilingual corpus, we extract tokens of varying character lengths ($1, 2, 3, 4, \dots$ characters) based on their frequency of occurrence.
These frequency-selected multilingual tokens are combined with the $64\text{K}$ Turkish tokens to form the final $128\text{K}$ vocabulary.
This custom hybrid construction ensures excellent morphological alignment for Turkish while preserving the teacher's multilingual capacity across $40+$ languages.

The choice of a $128\text{K}$ vocabulary size represents a key design trade-off. A smaller vocabulary, such as the $64\text{K}$ vocabulary used in the predecessor model \emph{embeddingmagibu-152m}\footnote{\url{https://huggingface.co/magibu/embeddingmagibu-152m}}, reduces the embedding table size but can lead to sequence truncation or high fragmentation for non-Turkish languages. With $131{,}072$ tokens and a hidden dimension of 768, the embedding table contains $131{,}072 \times 768 \approx 100.6\text{M}$ parameters. This saves approximately $96\text{M}$ parameters compared to the teacher's original $256\text{K} \times 768 \approx 196.6\text{M}$ embedding table, contributing to the student's reduced parameter footprint ($200\text{M}$ parameters). The impact of this vocabulary size choice is evaluated in Section~\ref{sec:ablations}.

\subsection{Weight-Preserving Cloning and Embedding Remapping}

The teacher embedding model is EmbeddingGemma with 300M parameters\footnote{\url{https://huggingface.co/google/embeddinggemma-300m}}~\cite{embeddinggemma_arxiv}. EmbeddingGemma is derived from the Gemma~3 architecture and produces 768-dimensional embedding vectors. It supports input sequences up to 2,048 tokens and includes prompt templates for query/document distinction in retrieval tasks.

The student model follows the SentenceTransformers format with a Gemma3TextModel backbone initialized from the teacher, mean pooling over token representations with \texttt{include\_prompt=True} for prompt-aware encoding, two linear projections $768 \rightarrow 3072 \rightarrow 768$ without bias terms or nonlinear activations (Identity), and final $\ell_2$ normalization to produce unit-length embedding vectors. The maximum sequence length is extended to 8,192 tokens. The final embedding dimension is 768, compatible with many downstream applications.

Changing the tokenizer fundamentally alters the vocabulary: the new multilingual tokenizer has different token identities than the teacher's original tokenizer. This makes the teacher's token embedding table incompatible with the student. The approach preserves transformer backbone weights (attention, feedforward, layer normalization) while recomputing a new embedding table through token-id mapping. For each token $j$ in the target vocabulary, the teacher tokenizer encoding of the same surface form is identified. This produces a mapping $\pi: j \mapsto (i_1, \dots, i_k)$, where $(i_1, \dots, i_k)$ is the sequence of teacher token IDs that corresponds to target token $j$. Given the mapping, the new embedding $E'_j$ for target token $j$ is initialized by combining the corresponding teacher embeddings:
\begin{equation}
E'_j = \text{Compose}(E_{i_1}, E_{i_2}, \dots, E_{i_k})
\end{equation}
where $E_i$ denotes the teacher embedding for token $i$. The composition strategy can be uniform averaging (MEAN), weighted averaging (WEIGHTED), or selection of a specific position (FIRST, LAST). Mean composition is used:
\begin{equation}
E'_j = \frac{1}{k}\sum_{m=1}^{k} E_{i_m}
\label{eq:embedding-remap-mean}
\end{equation}

This initialization avoids random token embeddings, reduces the embedding-table parameter count when moving from the teacher's 256K vocabulary to the student's 128K vocabulary, and preserves the transformer backbone weights exactly. The cloning procedure is implemented in the \texttt{transformer-cloner} package.

\subsection{Precomputed Distillation Dataset}

Running the teacher model at every training step is computationally expensive. To enable efficient training, teacher embeddings are precomputed for the training corpus and stored as a Hugging Face dataset.

The distillation corpus is built from a multilingual Wikipedia dataset covering 40 languages\footnote{\url{https://huggingface.co/datasets/alibayram/wikipedia-40-langs-with-embeddings}}. To prevent high-resource languages from dominating, a language-based quota is applied: Turkish and English are capped at 100K training examples each, while the other 38 languages are capped at 10K examples each. This results in a balanced corpus of approximately 580K training rows.

Teacher embeddings are generated using EmbeddingGemma-300M~\cite{embeddinggemma_arxiv} on the text fields of this corpus. Both the final normalized representations (\texttt{teacher\_embedding\_final}) and the representations prior to the dense projection layer (\texttt{teacher\_embedding\_pre\_dense}) are extracted, storing the final dataset in Parquet format on Hugging Face Hub.

\subsection{Offline Embedding Distillation}

The student is trained to match the teacher's embedding space using a cosine similarity objective. Let $t_i \in \mathbb{R}^d$ be the precomputed teacher embedding and $s_i \in \mathbb{R}^d$ the student embedding for input $x_i$. Using $\ell_2$-normalized embeddings $\hat{t}_i = t_i/\lVert t_i\rVert_2$ and $\hat{s}_i = s_i/\lVert s_i\rVert_2$, the cosine distillation loss is:
\begin{equation}
\mathcal{L}_{\text{cos}} = \frac{1}{N}\sum_{i=1}^{N} \left(1 - \hat{s}_i^\top \hat{t}_i\right)
\label{eq:cosine_loss}
\end{equation}
This loss is minimized when student and teacher embeddings are perfectly aligned (cosine similarity of 1) and maximized when they are orthogonal (cosine similarity of 0).

The distillation training uses the following hyperparameters, as specified in the Hugging Face model card: the final teacher embeddings (\texttt{teacher\_embedding\_final}) are distilled using the cosine loss in Equation~\ref{eq:cosine_loss} for one epoch with batch size 256 and learning rate $5 \times 10^{-5}$. A warmup ratio of 0.01, weight decay of 0.01, maximum gradient norm 1.0, and bf16 precision are used, with gradient checkpointing and \texttt{torch.compile} enabled. Checkpoints are saved every 100 steps.

Training is executed on a single NVIDIA A100 80GB GPU. The complete distillation process takes approximately four hours, demonstrating the efficiency of offline distillation compared to online approaches that would require running both teacher and student at each step.

\begin{figure}[!htbp]
\centering
\begin{minipage}{0.48\textwidth}
\centering
\includegraphics[width=\linewidth]{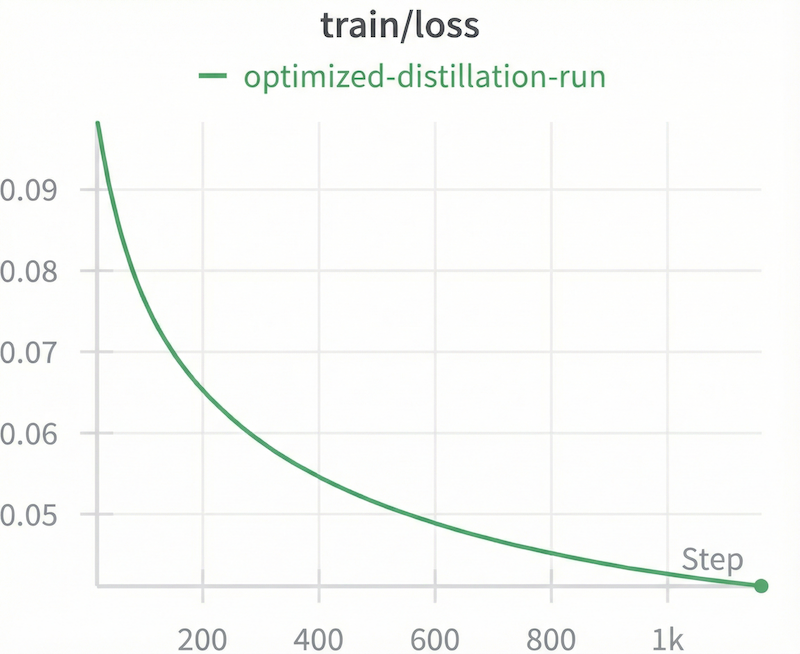}
\end{minipage}
\hfill
\begin{minipage}{0.48\textwidth}
\centering
\includegraphics[width=\linewidth]{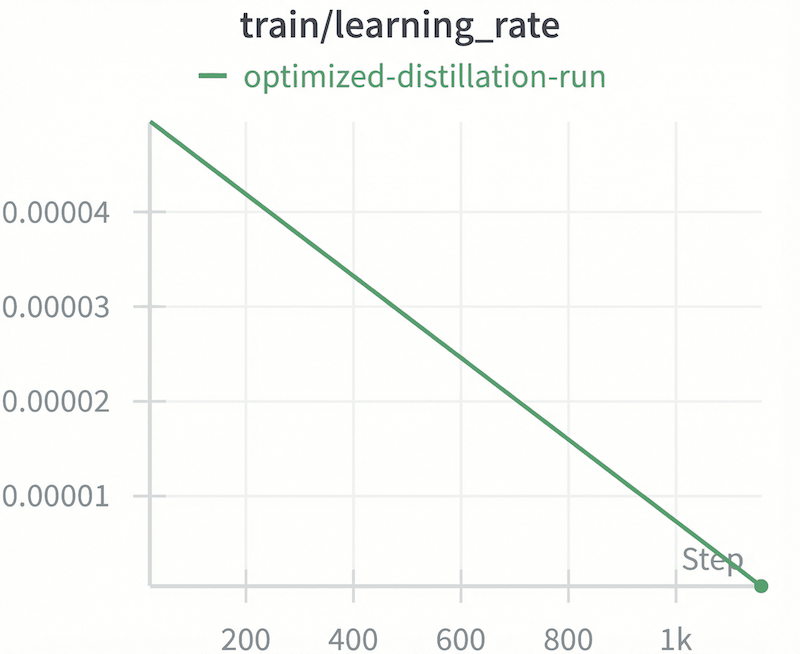}
\end{minipage}
\caption{Training curves from the offline distillation run. Left: cosine distillation loss decreasing from $\sim$0.09 to $\sim$0.05 over training. Right: learning rate schedule with warmup followed by cosine decay from $5 \times 10^{-5}$ to zero.}
\label{fig:training_curves}
\end{figure}

Training progress is tracked using Weights \& Biases\footnote{\url{https://api.wandb.ai/links/alibayram-ytu/srxzzhof}}. Figure~\ref{fig:training_curves} shows the training loss and learning rate curves from the distillation run.

The training logs indicate rapid early optimization (loss drops from 0.09 to 0.07 within the first 200 steps) followed by steady convergence to approximately 0.05 by the end of training. This rapid optimization confirms the viability of the mean-composition initialization, which aligns the vocabulary spaces before training begins.

\section{Experiments}
\label{sec:experiments}

Datasets include the Cosmos Turkish Corpus for tokenizer training, a balanced 40-language Wikipedia corpus with precomputed teacher embeddings ($\approx$580K rows) for distillation, and two evaluation benchmarks: STSbTR\footnote{\url{https://huggingface.co/datasets/figenfikri/stsb_tr}} and TR-MTEB~\cite{trmteb_paper}.

\paragraph{STSbTR.}
STSbTR is a translation-based adaptation of the English STS Benchmark containing sentence pairs rated on a semantic similarity scale from 0.0 to 5.0. The corpus contains 5,749 training pairs and 1,379 test pairs. Cosine similarity between sentence embeddings is computed, and both Pearson and Spearman correlation coefficients are reported on both splits.

\paragraph{TR-MTEB.}
TR-MTEB~\cite{trmteb_paper} is a comprehensive multi-task embedding benchmark containing 26 tasks across 7 categories: Retrieval (6 tasks), Classification (8 tasks), Clustering (2 tasks), STS (1 task), NLI (3 tasks), Bitext Mining (1 task), and Reranking (5 tasks). The macro-averaged score across categories and individual category averages are reported. An interactive benchmark results explorer is available online.

\paragraph{Baselines.}
Representative multilingual and Turkish-focused embedding models are included as baselines, including EmbeddingGemma-300M (teacher)~\cite{embeddinggemma_arxiv}, \emph{embeddingmagibu-152m} (predecessor), Multilingual E5 variants~\cite{e5_arxiv}, turkish-e5-large\footnote{\url{https://huggingface.co/ytu-ce-cosmos/turkish-e5-large}}, and TabiBERT~\cite{tabibert_arxiv}.

\section{Results and Discussion}
\label{sec:results}

\subsection{STSbTR Results}

Table~\ref{tab:embeddingmagibu-stsbtr} presents the evaluation of \emph{embeddingmagibu-200m} on STSbTR alongside 20 baselines.

\begin{table}[!htbp]
\centering
\caption{Semantic Textual Similarity (STSbTR) results for 21 models. Pearson (P) and Spearman (S) correlations are reported as percentages. Models are ranked by test Pearson correlation.}
\label{tab:embeddingmagibu-stsbtr}
\small
\begin{tabular}{clcccc}
\toprule
\textbf{Rank} & \textbf{Model} & \textbf{Test P} & \textbf{Test S} & \textbf{Train P} & \textbf{Train S} \\
\midrule
1 & emrecan/bert-base-tr-nli-stsb-tr & 83.45 & 83.13 & 93.79 & 92.31 \\
2 & intfloat/multiling.-e5-large-instruct & 80.50 & 81.23 & 82.75 & 81.29 \\
3 & ytu-ce-cosmos/turkish-e5-large & 79.73 & 79.99 & 80.90 & 79.06 \\
4 & intfloat/multilingual-e5-base & 78.61 & 78.35 & 79.86 & 77.38 \\
5 & selmanbaysan/tr-emb-fine-tuned & 78.08 & 78.35 & 82.15 & 80.61 \\
\textbf{6} & \textbf{embeddingmagibu-200m (Ours)} & \textbf{77.55} & \textbf{77.45} & \textbf{82.35} & \textbf{80.27} \\
7 & intfloat/multilingual-e5-small & 76.50 & 76.15 & 79.10 & 76.88 \\
8 & newmindai/Mursit-Large-TR-Retr. & 76.42 & 74.60 & 76.33 & 72.58 \\
9 & trmteb/turkish-embedding-model & 76.05 & 74.94 & 77.07 & 74.87 \\
10 & newmindai/Mursit-Base-TR-Retr. & 76.02 & 74.05 & 74.47 & 70.72 \\
11 & embeddinggemma-300m (Teacher) & 73.84 & 72.92 & 73.91 & 71.94 \\
\textbf{12} & \textbf{embeddingmagibu-152m} & \textbf{72.92} & \textbf{71.84} & \textbf{73.26} & \textbf{70.24} \\
13 & dbmdz/bert-base-turkish-cased & 57.32 & 56.72 & 59.37 & 57.02 \\
\bottomrule
\end{tabular}
\end{table}

\emph{embeddingmagibu-200m} ranks 6th out of 21 models, achieving 77.55\% Pearson correlation on the test set. The student model significantly outperforms its teacher, EmbeddingGemma-300M (73.84\% test Pearson, +3.71\% absolute), and its predecessor, \emph{embeddingmagibu-152m} (72.92\% test Pearson, +4.63\% absolute). On the training set, \emph{embeddingmagibu-200m} ranks 3rd overall with 82.35\% Pearson correlation. These results confirm that tokenizer surgery combined with offline distillation successfully transfers the teacher's semantic capabilities to the Turkish target space.

\subsection{TR-MTEB Results}

Table~\ref{tab:embeddingmagibu-trmteb} summarizes the macro-averaged results across TR-MTEB categories.

\begin{table}[!htbp]
\centering
\caption{TR-MTEB macro-averaged category results for selected models. Rank is out of 26 models evaluated. Category averages cover Retrieval (6 tasks), Classification (8 tasks), Clustering (2 tasks), STS (1 task), NLI (3 tasks), and Bitext Mining (1 task).}
\label{tab:embeddingmagibu-trmteb}
\scriptsize
\resizebox{\linewidth}{!}{
\begin{tabular}{lcccccccc}
\toprule
\textbf{Model} & \textbf{Rank} & \textbf{Avg} & \textbf{Retr.} & \textbf{Cls.} & \textbf{Clust.} & \textbf{STS} & \textbf{NLI} & \textbf{Bitext} \\
\midrule
text-embedding-3-small & 1/26 & 66.5 & 78.1 & 69.5 & 62.1 & 70.8 & 57.2 & 91.6 \\
ytu-ce-cosmos/turkish-e5-large & 2/26 & 66.0 & 77.0 & 72.6 & 60.7 & 80.0 & 62.7 & 99.2 \\
intfloat/multiling.-e5-large-instruct & 3/26 & 65.9 & 76.1 & 73.7 & 62.1 & 81.2 & 63.0 & 99.0 \\
embeddinggemma-300m (Teacher) & 4/26 & 65.2 & 75.9 & 71.8 & 62.4 & 72.9 & 60.6 & 96.8 \\
alibaba-NLP/gte-multiling.-base & 5/26 & 64.4 & 75.2 & 67.8 & 60.4 & 80.7 & 66.7 & 98.0 \\
\textbf{embeddingmagibu-200m (Ours)} & \textbf{7/26} & \textbf{63.9} & \textbf{72.2} & \textbf{68.5} & \textbf{61.4} & \textbf{77.5} & \textbf{67.9} & \textbf{97.0} \\
selmanbaysan/tr-emb-fine-tuned & 8/26 & 63.4 & 71.4 & 70.9 & 61.5 & 78.5 & 70.8 & 86.9 \\
\textbf{embeddingmagibu-152m} & \textbf{12/26} & \textbf{60.2} & \textbf{71.3} & \textbf{65.5} & \textbf{61.6} & \textbf{71.8} & \textbf{55.6} & \textbf{90.1} \\
trmteb/turkish-embedding-model & 14/26 & 59.8 & 69.5 & 70.7 & 63.6 & 75.1 & 56.9 & 37.8 \\
dbmdz/bert-base-turkish-uncased & 17/26 & 49.5 & 53.6 & 70.1 & 60.7 & 56.8 & 52.4 & 7.1 \\
KocLab-Bilkent/BERTurk-Legal & 22/26 & 39.3 & 24.6 & 64.8 & 59.1 & 42.3 & 49.5 & 1.6 \\
\bottomrule
\end{tabular}
}
\end{table}

On TR-MTEB, \emph{embeddingmagibu-200m} achieves an average score of 63.9\%, ranking 7th out of 26 models. Compared to its teacher (EmbeddingGemma-300M, 65.2\%), the student model achieves competitive results despite a 33\% reduction in parameters. In task-level comparisons, \emph{embeddingmagibu-200m} outperforms its teacher on STS (77.5\% vs.\ 72.9\%), NLI (67.9\% vs.\ 60.6\%), and Bitext Mining (97.0\% vs.\ 96.8\%). It lags on Retrieval (72.2\% vs.\ 75.9\%) and Classification (68.5\% vs.\ 71.8\%), suggesting that adapting the tokenizer preserves semantic compositionality while introducing minor tradeoffs in classification, which can be mitigated by task-specific fine-tuning.

\subsection{Discussion}

\paragraph{Morphological Alignment and Learning Dynamics.}
Agglutinative languages pose challenges for tokenization due to word-form sparsity. Our results show that resolving fragmentation by using a larger, Turkish-optimized vocabulary yields downstream benefits. In tasks requiring fine-grained semantic compositionality (STS and NLI), \emph{embeddingmagibu-200m} achieves absolute gains over its teacher (+4.6\% and +7.3\% respectively). Composing embeddings of morphologically aligned subwords provides a more stable semantic foundation, allowing the student to converge quickly during distillation.

\paragraph{Long Context and RAG Suitability.}
Most monolingual Turkish models are constrained by a 512-token context window. This limitation forces RAG pipelines to use small text chunks, fragmenting document structure. With an 8K-token context window and a reduced token footprint, \emph{embeddingmagibu-200m} can encode whole documents without chunking. The token footprint reduction achieved by the target tokenizer further extends the effective context window, making the model suitable for enterprise RAG applications and document retrieval in Turkish.

\paragraph{Cost--Quality Frontier.}
The total training cost of \$5--\$20 and approximately four GPU hours places \emph{embeddingmagibu-200m} at an unprecedented cost--quality point. It achieves 98.0\% of the teacher's TR-MTEB average while using 33\% fewer parameters and outperforms the teacher on the STS, NLI, and Bitext Mining categories. This confirms that language-specific tokenizer adaptation with targeted distillation can match or exceed a larger multilingual model's performance on target-language benchmarks.

\section{Ablations and Analysis}
\label{sec:ablations}

\subsection{Vocabulary Size Ablation: 64K vs.\ 128K}

The vocabulary size directly controls the trade-off between model parameters and downstream representation quality. We compare \emph{embeddingmagibu-200m} (128K vocabulary) with its predecessor \emph{embeddingmagibu-152m} (64K vocabulary). Both models share the same EmbeddingGemma backbone and distillation recipe, differing only in vocabulary size, embedding layer footprint, and distillation corpus.

Table~\ref{tab:embeddingmagibu-200m-vs-152m} presents the comparison across TR-MTEB categories.

\begin{table}[!htbp]
\centering
\caption{Comparison of \emph{embeddingmagibu-200m} (128K vocabulary) and \emph{embeddingmagibu-152m} (64K vocabulary) across TR-MTEB tasks.}
\label{tab:embeddingmagibu-200m-vs-152m}
\small
\begin{tabular}{lccc}
\toprule
\textbf{Category} & \textbf{200m (128K)} & \textbf{152m (64K)} & \textbf{$\Delta$} \\
\midrule
Overall Average (26 tasks) & 63.9 & 60.2 & +3.7 \\
Overall Rank (out of 26) & 7/26 & 12/26 & +5 ranks \\
\midrule
Retrieval & 72.2 & 71.3 & +0.9 \\
Classification & 68.5 & 65.5 & +3.0 \\
Clustering & 61.4 & 61.6 & $-$0.2 \\
STS & 77.5 & 71.8 & +5.7 \\
NLI & 67.9 & 55.6 & +12.3 \\
Bitext Mining & 97.0 & 90.1 & +6.9 \\
\bottomrule
\end{tabular}
\end{table}

Doubling the vocabulary from 64K to 128K increases the embedding layer from 49.5M to 100.6M parameters (+51M). This parameter increase yields substantial performance gains (+3.7\% overall). The improvement is most pronounced in NLI (+12.3\% absolute), STS (+5.7\%), and Bitext Mining (+6.9\%). The clustering category shows negligible change ($-$0.2\%). These results indicate that expanding the vocabulary improves representational compositionality, particularly in semantic reasoning and textual similarity tasks.

\subsection{Scale--Performance Frontier}

To understand the efficiency of cross-lingual transfer, we analyze the scale--performance frontier across the 152M student, 200M student, and 300M teacher:

\begin{itemize}
    \item \textbf{EmbeddingGemma-300M (Teacher):} 300M parameters (196.6M in embeddings), 256K vocabulary. TR-MTEB average: 65.2\%.
    \item \textbf{embeddingmagibu-200m (Ours):} $\approx$205M parameters (100.6M in embeddings), 128K vocabulary. TR-MTEB average: 63.9\%.
    \item \textbf{embeddingmagibu-152m:} $\approx$154M parameters (49.5M in embeddings), 64K vocabulary. TR-MTEB average: 60.2\%.
\end{itemize}

\emph{embeddingmagibu-200m} achieves 98.0\% of the teacher's average performance while using 33\% fewer parameters, and outperforms the teacher on STS (77.5\% vs.\ 72.9\%) and NLI (67.9\% vs.\ 60.6\%). This confirms that language-specific tokenizer adaptation with targeted distillation can match or exceed a larger multilingual model's performance on target-language benchmarks while reducing the parameter footprint.

\subsection{Parameter Footprint}

Table~\ref{tab:parameter-comparison} compares the parameter sizes of key models.

\begin{table}[!htbp]
\centering
\caption{Parameter footprint comparison. Vocabulary size, embedding layer parameters, backbone parameters, and total parameters.}
\label{tab:parameter-comparison}
\small
\begin{tabular}{lcccc}
\toprule
\textbf{Model} & \textbf{Vocab} & \textbf{Emb.\ Params} & \textbf{Backbone} & \textbf{Total} \\
\midrule
multiling.-e5-large-instruct & 250K & 256.0M & 304.0M & 560.0M \\
embeddinggemma-300m & 256K & 196.6M & 104.0M & 300.6M \\
\textbf{embeddingmagibu-200m} & \textbf{128K} & \textbf{100.6M} & \textbf{104.0M} & \textbf{$\approx$205M} \\
embeddingmagibu-152m & 64K & 49.5M & 104.0M & $\approx$154M \\
\bottomrule
\end{tabular}
\end{table}

By trimming the vocabulary from 256K to 128K, the embedding layer parameters are reduced by 48.8\% (196.6M to 100.6M). This directly translates to a smaller disk footprint, lower GPU VRAM consumption, and faster inference throughput.

\subsection{Controlled Tokenizer Ablations}

To isolate the role of tokenization design from other confounders (pretraining history, data mix), we refer to the controlled tokenizer experiments by Bayram et al.~\cite{hybrid_tokenization_arxiv}. In those experiments, four different tokenizers (MFT, Tabi, Cosmos, and Mursit) were compared using the same model backbone (EmbeddingGemma-300M encoder), same training corpus, and identical random initialization (seed=42). All four students were trained with cosine distillation on the same teacher embeddings. The results show that tokenizers trained on large monolingual corpora achieve higher morphological alignment and lower subword fragmentation, which correlates with downstream performance on STS, NLI, and retrieval tasks. This finding motivated the choice of training a large-vocabulary SentencePiece-BPE tokenizer on the Cosmos Turkish Corpus.

\section{Limitations}
\label{sec:limitations}

Several limitations should be noted. The student model is bounded by the semantic space of the teacher; any representation errors or biases present in EmbeddingGemma-300M may be transferred to the student. By reducing the vocabulary from 256K to 128K, the model's capacity to represent non-target languages is reduced; while it retains representation quality for Turkish and English (included in the distillation corpus), performance on lower-resource non-target languages may degrade compared to the teacher. The pooling and projection layers are linear, preserving the geometry of the teacher's space; non-linear projections or task-specific fine-tuning could improve performance on classification or clustering tasks. The mean-composition initialization ignores polysemy and context: a surface-form token may map to teacher subwords whose embeddings reflect multiple senses, and averaging cannot disambiguate which sense is relevant. Although the model supports 8,192-token inputs, performance on very long documents is not extensively evaluated; such content may require chunking and aggregation strategies beyond simple mean pooling. The training recipe focuses on matching teacher embeddings for single texts and does not include supervised contrastive training on Turkish NLI/IR data, which may limit ceiling performance. Finally, while high-level training procedures and artifacts are released, some implementation details (e.g., exact random seeds, data shuffling) may affect exact reproducibility.

\section{Reproducibility}
\label{sec:reproducibility}

Comprehensive information is provided to facilitate reproduction of results. The model weights are available on Hugging Face, and for local deployment the model is also distributed via Ollama. The precomputed distillation dataset is released as a Hugging Face dataset. Released artifacts include tokenizer files (\texttt{tokenizer.model}, \texttt{tokenizer.json}, \texttt{tokenizer\_config.json}), model configuration (\texttt{config.json}, \texttt{modules.json}), module configurations and weights (\texttt{1\_Pooling/}, \texttt{2\_Dense/}, \texttt{3\_Dense/}), and the model weights (\texttt{model.safetensors}). Training logs are available via Weights \& Biases.

The following packages are required:
\begin{verbatim}
pip install -U sentence-transformers datasets sentencepiece
pip install -U transformer-cloner distil-trainer
\end{verbatim}

Pipeline steps are structured as follows: (1)~hybrid tokenizer construction via frequency selection and token pruning, (2)~weight-preserving teacher model cloning with embedding remapping using the \texttt{transformer-cloner} package, (3)~teacher embeddings generation, and (4)~offline student model distillation using the \texttt{distil-trainer} package. The complete conceptual Python code for the custom tokenizer construction and the training scripts are provided in Appendix~\ref{sec:appendix_code}.

Hardware requirements include a single NVIDIA A100 80GB GPU (or an equivalent GPU with $\geq$40GB VRAM). Distillation takes approximately 4 hours, and the precomputed-embeddings dataset requires approximately 5--10GB of storage. Complete training hyperparameters as specified in the Hugging Face model card:
\begin{center}
\small
\begin{tabular}{ll}
\toprule
Hyperparameter & Value \\
\midrule
Epochs & 1 \\
Batch size & 256 \\
Learning rate & $5 \times 10^{-5}$ \\
Warmup ratio & 0.01 \\
Weight decay & 0.01 \\
Max gradient norm & 1.0 \\
Precision & bf16 \\
Gradient checkpointing & Enabled \\
torch.compile & Enabled \\
Loss function & Cosine \\
Target type & Final embeddings \\
\bottomrule
\end{tabular}
\end{center}

\section{Conclusion}
\label{sec:conclusion}

\emph{embeddingmagibu-200m} is a Turkish-focused sentence embedding model with an extended 8,192-token context window, a 128K multilingual vocabulary, and 768-dimensional outputs. The three-stage pipeline---hybrid tokenizer construction, weight-preserving model cloning with embedding remapping, and offline distillation from precomputed teacher embeddings---provides an efficient approach to language-specific model adaptation.

Empirically, the student surpasses its EmbeddingGemma teacher on STSbTR (77.55\%/77.45\% vs.\ 73.84\%/72.92\% Pearson/Spearman), indicating that Turkish-optimized tokenization combined with distillation can improve performance over a multilingual teacher on Turkish semantic similarity. On TR-MTEB (26 tasks), the model achieves an average score of 63.9\% (7th out of 26 models) while using approximately 200M parameters---a 33\% reduction from the teacher. The total training cost of \$5--\$20 and approximately four hours of GPU time demonstrates the accessibility of the approach.

Future work includes comparing alternative tokenizer adaptation methods (e.g., WECHSEL~\cite{wechsel_naacl2022}, hybrid tokenization~\cite{hybrid_tokenization_arxiv}); extending the pipeline to other morphologically rich languages (such as Finnish, Hungarian, and Korean); applying non-linear projection layers to improve classification and clustering performance; and integration with retrieval-augmented generation pipelines for long-context Turkish applications.

In addition to the model, supporting infrastructure is released for adoption and benchmarking, including the open-source cloning and distillation tools, precomputed embedding datasets for offline training, and the Hugging Face Space for interactive exploration and benchmark inspection.

\bibliographystyle{plainnat}
\bibliography{references}

\appendix
\section{Implementation Code}
\label{sec:appendix_code}

This appendix provides the Python implementation snippets for the custom tokenizer construction, model cloning, teacher embedding generation, and student training.

\subsection{Conceptual Custom Tokenizer Construction}
The tokenizer was constructed using a custom pipeline that selects high-frequency Turkish tokens, prunes alternative teacher tokenizer representations, and merges them with frequency-filtered multilingual tokens from the Wikipedia-40-langs corpus. Below is a conceptual implementation of this hybrid process:

\begin{verbatim}
import sentencepiece as spm
from transformers import AutoTokenizer

# 1. Train a temporary Turkish tokenizer to analyze subword frequencies
spm.SentencePieceTrainer.train(
    input='cosmos_turkish_corpus.txt',
    model_prefix='turkish_bpe_raw',
    vocab_size=100000,
    model_type='bpe'
)

# Load raw vocab and extract top 64K (65,536) frequent tokens
turkish_vocab = load_and_sort_by_frequency('turkish_bpe_raw.vocab')
turkish_64k_tokens = turkish_vocab[:65536]

# 2. Prune redundant teacher (Gemma) tokens that can be
# resolved by these 64K Turkish tokens
teacher_tokenizer = AutoTokenizer.from_pretrained(
    'google/embeddinggemma-300m'
)
pruned_teacher_vocab = prune_redundant_tokens(
    teacher_tokenizer, turkish_64k_tokens
)

# 3. Perform frequency analysis on the Wikipedia 40-languages dataset
# Select multilingual tokens of lengths 1, 2, 3, 4,... by usage frequency
wikipedia_tokens = analyze_multilingual_frequencies(
    dataset_path='alibayram/wikipedia-40-langs',
    max_token_lengths=[1, 2, 3, 4]
)

# 4. Merge selections to build the final 128K (131,072) tokenizer
final_128k_vocab = combine_vocabularies(
    turkish_tokens=turkish_64k_tokens,
    multilingual_tokens=wikipedia_tokens,
    target_size=131072
)

# Export the new multilingual tokenizer model
save_custom_tokenizer(
    final_128k_vocab, 'embeddingmagibu_200m_tokenizer.model'
)
\end{verbatim}

\subsection{Weight-Preserving Model Cloning}
Once the new tokenizer model is saved, the student model is initialized by cloning the teacher model (\texttt{embeddinggemma-300m}) weights and remapping the embedding table via the \texttt{transformer-cloner} package:

\begin{verbatim}
from transformer_cloner import TransformerCloner

cloner = TransformerCloner(
    source_model='google/embeddinggemma-300m',
    target_tokenizer='./embeddingmagibu_200m_tokenizer.model'
)
cloner.clone(output_path='./cloned_model')
\end{verbatim}

\subsection{Teacher Embedding Generation}
Before beginning training, the teacher's embeddings are precomputed over the multilingual Wikipedia corpus to enable efficient offline distillation using the \texttt{distil-trainer} package:

\begin{verbatim}
from distil_trainer.data import TeacherEmbeddingsGenerator

generator = TeacherEmbeddingsGenerator(
    teacher_model='google/embeddinggemma-300m'
)
generator.generate(
    dataset='wikipedia_40_langs',
    output_path='./embeddings_dataset'
)
\end{verbatim}

\subsection{Offline Embedding Distillation}
Finally, the student model is trained to minimize the cosine distance to the precomputed teacher embeddings using the distillation trainer:

\begin{verbatim}
from distil_trainer import EmbeddingDistillationTrainer

trainer = EmbeddingDistillationTrainer(
    student_model='./cloned_model',
    embeddings_dataset='./embeddings_dataset',
    target_type='final',
    loss='cosine',
    batch_size=256,
    learning_rate=5e-5,
    num_epochs=1,
    precision='bf16'
)
trainer.train()
\end{verbatim}

\end{document}